\documentclass{ecai}  % use option [doubleblind] for double blind submission

\usepackage{graphicx}
\usepackage{latexsym}
\usepackage{amsmath}
\usepackage{booktabs} % For better tables

\begin{document}

\begin{frontmatter}

\title{SpatialText: A Pure-Text Cognitive Benchmark for Spatial Understanding in Large Language Models}

\author[A]{\fnms{Peiyao}~\snm{Jiang}\thanks{Email: 3220100997@zju.edu.cn}}
\author[B]{\fnms{Zequn}~\snm{Qin}\thanks{Email: zequnqin@zju.edu.cn}}
\author[A]{\fnms{Xi}~\snm{Li}\thanks{Email: xilizju@zju.edu.cn}}

\address[A]{Zhejiang University}
\address[B]{School of Software Technology, Zhejiang University}

\begin{abstract}
Genuine spatial reasoning relies on the capacity to construct and manipulate coherent internal spatial representations, often conceptualized as mental models, rather than merely processing surface linguistic associations. While large language models exhibit advanced capabilities across various domains, existing benchmarks fail to isolate this intrinsic spatial cognition from statistical language heuristics. Furthermore, multimodal evaluations frequently conflate genuine spatial reasoning with visual perception. To systematically investigate whether models construct flexible spatial mental models, we introduce SpatialText, a theory-driven diagnostic framework. Rather than functioning simply as a dataset, SpatialText isolates text-based spatial reasoning through a dual-source methodology. It integrates human-annotated descriptions of real 3D indoor environments—which capture natural ambiguities, perspective shifts, and functional relations—with code-generated, logically precise scenes designed to probe formal spatial deduction and epistemic boundaries. Systematic evaluation across state-of-the-art models reveals fundamental representational limitations. Although models demonstrate proficiency in retrieving explicit spatial facts and operating within global, allocentric coordinate systems, they exhibit critical failures in egocentric perspective transformation and local reference frame reasoning. These systematic errors provide strong evidence that current models rely heavily on linguistic co-occurrence heuristics rather than constructing coherent, verifiable internal spatial representations. SpatialText thus serves as a rigorous instrument for diagnosing the cognitive boundaries of artificial spatial intelligence.
\end{abstract}

\end{frontmatter}

\begin{figure*}[t]
    \centering
    % width=0.9\textwidth 表示图片宽度占正文宽度的 90%
    \includegraphics[width=1\textwidth]{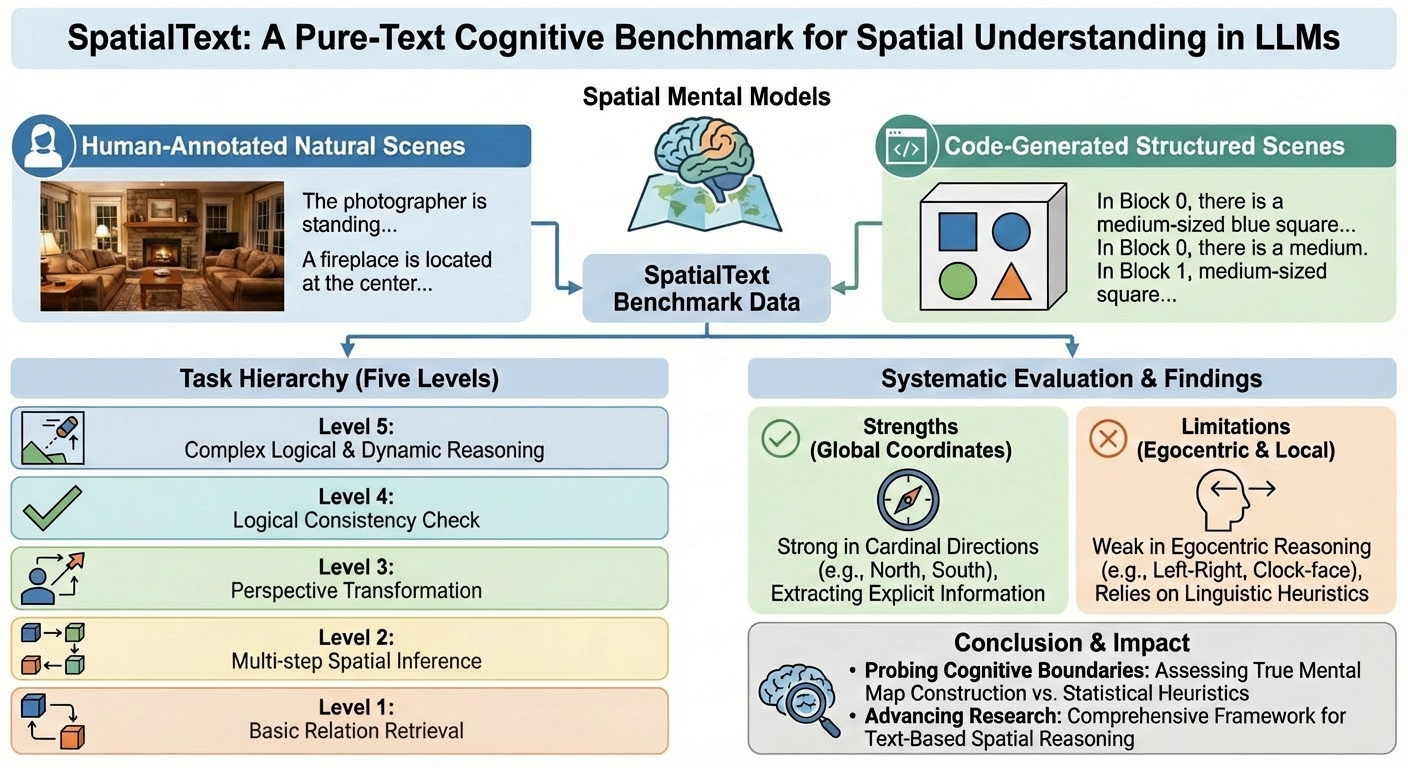} 
    
    \caption{Overview of the SpatialText Framework: A dual-source diagnostic benchmark for evaluating text-based spatial cognition in Large Language Models (LLMs). The framework constructs datasets through a dual-source methodology (human-annotated real scenes and code-generated logical scenes) and encompasses five core spatial tasks ranging from localization to mental rotation, aiming to reveal the models' capacity for mental model construction through multi-dimensional evaluation.}
    \label{fig:human_annotation_examples}
\end{figure*}

\section{Introduction}
Spatial reasoning is a fundamental component of human intelligence and a prerequisite for meaningful interaction with the physical world. Humans possess a remarkable ability to construct internal spatial representations—often referred to as mental maps—from purely linguistic input. Cognitive psychology has long characterized this capability through the theory of Spatial Mental Models \cite{tversky1991}, which posits that language comprehension induces an internal, viewpoint-independent spatial structure that supports flexible reasoning and perspective transformation.

In the transition toward artificial general intelligence, a critical question arises: do Large Language Models (LLMs) genuinely possess the capacity to ground language in such internal spatial manifolds, or do they merely simulate reasoning through sophisticated surface-level linguistic heuristics? This question remains unresolved because existing evaluation frameworks are ill-equipped to diagnose representational depth. Widely used benchmarks such as GSM8K \cite{gsm8} and MMLU \cite{Hendrycks2020MeasuringMM} evaluate general reasoning but do not isolate spatial cognition as an independent mental faculty. Multimodal benchmarks incorporating visual inputs conflate spatial reasoning with perception—bypassing the core cognitive challenge of constructing a "mental map" without direct sensory input. Meanwhile, existing text-based spatial benchmarks rely on simplified 2D grid worlds that lack real-world complexity, or inadvertently permit solutions through statistical pattern matching rather than genuine spatial deduction. For instance, a model might correctly place a "bed" near a "wall" not because it has mapped the room's geometry, but because those terms frequently co-occur in its training corpus.

To bridge this gap, we introduce SpatialText, a theory-driven diagnostic framework designed to isolate and probe the intrinsic spatial cognition of LLMs. Moving beyond simple dataset construction, SpatialText is engineered to test the "mental model hypothesis" through a dual-source, complementary data strategy. The primary component consists of human-annotated descriptions of real 3D indoor scenes, capturing the pragmatic ambiguity, functional relations, and perspective shifts characteristic of natural spatial language. This is complemented by code-generated, logically structured environments that serve as a rational scaffold—enabling precise evaluation of formal properties such as transitivity, coordinate transformation, and epistemic boundary recognition in both omniscient and non-omniscient scenarios. Together, these sources balance ecological validity with logical rigor.

Based on this foundation, SpatialText defines a hierarchy of reasoning tasks ranging from basic relation retrieval to perspective transformation, global consistency checking, and counterfactual inference. Through systematic evaluation of state-of-the-art models, we uncover a profound disconnect between linguistic fluency and representational grounding. While models excel at retrieving explicit spatial facts and reasoning within global (allocentric) frames, they exhibit systematic and catastrophic failures in egocentric transformations. These persistent errors—such as the "bed-north hallucination" where models default to high-frequency spatial associations—suggest that current LLMs do not construct verifiable internal spatial models. Instead, their "reasoning" collapses when tasks demand stable geometric manifolds, revealing reliance on statistical heuristics rather than coherent mental representations.

In summary, our contributions are threefold. First, we propose SpatialText, a carefully designed text-only benchmark for evaluating spatial cognition in large language models. Second, we introduce a dual-source data construction paradigm that jointly captures the ambiguity of real-world spatial language and the precision of formal spatial logic. Third, through systematic evaluation, we reveal previously underexplored cognitive boundaries of current models, providing rigorous evidence that representational grounding—not linguistic fluency—remains the primary bottleneck in achieving embodied artificial intelligence.

\section{Related Work}
\paragraph{Spatial Mental Model}
Theoretical frameworks in cognitive psychology suggest that spatial language comprehension is not merely symbolic manipulation, but a process of constructing viewpoint-flexible internal representations known as \emph{spatial mental models} \cite{johnsonlaird1983mental, tversky1991}. These models enable humans to decouple spatial information from specific linguistic descriptions, supporting mental rotation, perspective transformation, and inference of non-explicit relations \cite{tversky2005}. While these theories define the essence of spatial intelligence, AI evaluation has largely failed to distinguish between genuine representational grounding and surface-level pattern matching. SpatialText operationalizes this cognitive requirement by evaluating whether models maintain a stable internal manifold across shifting reference frames.

\paragraph{Lack of Benchmarks for Pure Spatial Reasoning.}
Mainstream LLM benchmarks such as GSM8K \cite{gsm8} and MMLU \cite{Hendrycks2020MeasuringMM} focus on mathematical deduction, symbolic logic, and world knowledge. Within these frameworks, spatial reasoning is treated as a marginal sub-type—often reduced to arithmetic manipulation—overlooking the unique geometric and topological constraints inherent in spatial cognition. 

\paragraph{Over-Idealized Spatial Descriptions.}
Prior text-based spatial reasoning benchmarks typically rely on simplified synthetic environments, such as 2D grid worlds (e.g., bAbI \cite{weston2015babai}, \textsc{StepGame} \cite{stepgame}) or symbolic graphs (e.g., FloorPlanQA \cite{floorplanqa}). While these controlled environments enable precise logic testing, they lack the topological complexity, functional nuances, and ambiguity of real-world 3D spaces. Models can often "solve" these benchmarks by exploiting linguistic co-occurrence or performing simple graph traversals that do not require a coherent spatial mental map. In contrast, SpatialText utilizes naturalistic descriptions of real-world scenes, forcing models to navigate the "noise" of functional relations and 3D occlusion that cannot be resolved through 2D symbolic logic alone.

\paragraph{Visual Inputs Obscure Textual Spatial Reasoning.}
Multimodal benchmarks including CLEVR \cite{johnson2017clevr}, GQA \cite{hudson2019gqa}, and Room-to-Room navigation \cite{anderson2018vision} evaluate spatial grounding through visual perception. However, these tasks conflate spatial reasoning with visual recognition; high performance may stem from extracting visual cues rather than constructing an internal model of the environment. By deliberately excluding visual input, SpatialText isolates the linguistic and cognitive mechanisms of spatial reasoning. This design ensures that success is contingent on the model's capacity to build an internal representation from text alone—addressing a diagnostic gap that multimodal evaluations leave unresolved.

\section{Data Construction}
\label{sec:data_construction}

Evaluating whether large language models can construct internal spatial representations from text involves multiple methodological considerations. Different data sources emphasize different aspects of spatial reasoning: natural scene descriptions reflect the ambiguity, pragmatic dependency, and perspective variability of real-world language use, whereas programmatically generated scenarios provide precise control over geometric structure and logical consistency. Rather than relying on a single data paradigm, we adopt a dual-source complementary design. By integrating human-annotated natural scenes with code-generated structured environments, we aim to capture both ecological richness and formal diagnosability, allowing these two perspectives to jointly inform the assessment of a shared cognitive target—text-based spatial model construction.

\subsection{Human-Annotated Natural Indoor Scenes}

The human-annotated portion of the benchmark targets spatial reasoning in naturally occurring environments. Unlike structured synthetic settings, real-world spatial descriptions are embedded in semantic ambiguity, functional context, and perspective-dependent expressions. These characteristics require models to interpret implicit constraints, resolve referential uncertainty, and maintain coherent spatial representations under realistic communicative conditions. By grounding evaluation in authentic indoor scenes, this component emphasizes robustness to linguistic variation and contextual nuance, assessing whether models can sustain spatial consistency in non-idealized settings.

\paragraph{Annotation Source: LSUN} 
To ground the human-annotated portion of the benchmark in realistic spatial environments, we source images from the LSUN indoor scene dataset. LSUN provides large-scale, high-diversity indoor scene images that capture natural object layouts, functional arrangements, and viewpoint variability. From this dataset, we select five common indoor categories: Bedroom, Living Room, Dining Room, Kitchen, and Classroom.

To ensure spatial richness while avoiding extreme complexity, we apply a two-stage filtering process. First, we exclude scenes that are overly sparse or visually cluttered. Second, we retain only images containing at least five salient objects with clearly identifiable spatial relations. This selection strategy balances structural complexity with annotation feasibility, ensuring that each scene supports multi-object relational reasoning without introducing unnecessary perceptual noise.

In total, 100 images are selected, with 20 images per category.

\paragraph{Annotation Strategy} 
To probe different dimensions of textual spatial reasoning, annotations are designed under three distinct reference-frame strategies: egocentric, allocentric, and hybrid. Rather than merely describing object locations, these strategies systematically manipulate how spatial relations are framed, thereby imposing different cognitive demands on the model.

Egocentric descriptions encode relations relative to an observer or the intrinsic orientation of an object (e.g., “to the left of the bed”). This strategy emphasizes local spatial reasoning and object-centered perspective alignment.

Allocentric descriptions rely on environment-centered anchors, such as cardinal directions or fixed global references. This requires maintaining a stable global spatial representation independent of observer position.

Hybrid descriptions combine absolute orientation cues with relative or clock-based expressions (e.g., “at the 3 o’clock direction on the east side”). This formulation introduces cross-reference-frame transformation, requiring the model to dynamically reconcile multiple coordinate systems within a single internal representation.

By distributing annotations across these three strategies, we ensure that the benchmark captures varied forms of spatial representation construction rather than a single descriptive style.

\begin{figure*}[t]
    \centering
    % width=0.9\textwidth 表示图片宽度占正文宽度的 90%
    \includegraphics[width=1\textwidth]{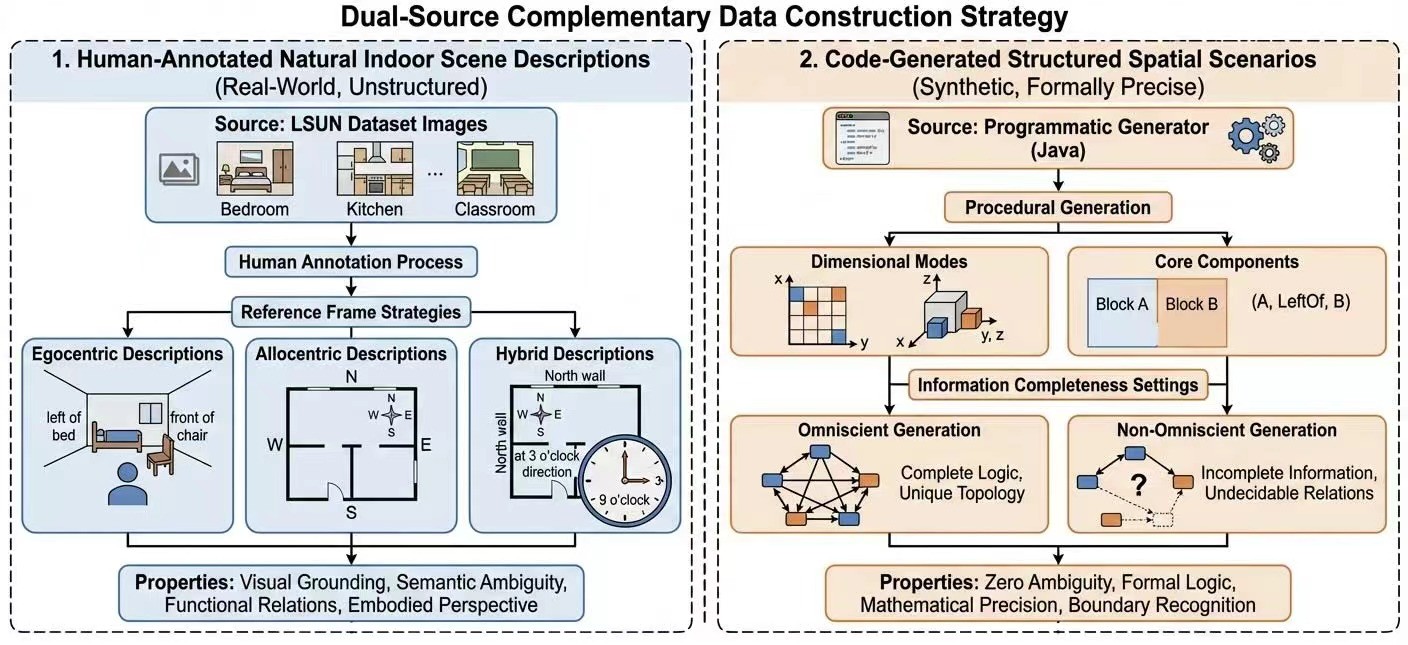} 
    
    \caption{Overview of the SpatialText Data Generation Framework. The framework integrates two complementary pipelines: (Left) Human-Annotated Real-world Scenes, which utilize indoor images from the LSUN dataset annotated via three progressive linguistic strategies ; (Right) Code-Generated Synthetic Scenes, provides diverse combinatorial descriptions across 2D/3D dimensions and Omniscient/Non-omniscient epistemic perspectives.}
    \label{fig:human_annotation_examples}
\end{figure*}

\subsection{Code-Generated Structured Scenes}

Complementing the natural data, the synthetic component isolates the formal structure of spatial reasoning under controlled conditions. Instead of reflecting communicative variability, programmatically generated scenes provide explicitly defined geometric relations and logically complete constraint systems. This controlled design eliminates semantic ambiguity and enables precise verification of relational consistency, coordinate transformation, and deductive inference. By varying dimensional complexity and information completeness, the synthetic dataset offers a diagnostic environment for examining the internal coherence and epistemic calibration of spatial reasoning processes.

\paragraph{Generation Settings} 
The synthetic dataset is constructed under a controlled generation framework that systematically varies spatial complexity and information availability. Specifically, we organize instances along two orthogonal dimensions: spatial dimensionality (2D vs. 3D) and epistemic completeness (omniscient vs. non-omniscient).

The 2D setting defines object positions within a planar coordinate system (x, y), emphasizing horizontal spatial relations such as left–right and front–back. The 3D setting extends this structure by introducing a vertical axis (z), thereby enabling reasoning over height, stacking, and volumetric containment. This dimensional variation increases structural complexity and requires models to maintain more elaborate internal representations.

Orthogonal to dimensionality, we manipulate information completeness. In the omniscient condition, the textual description provides logically sufficient constraints to reconstruct a unique global topology. In contrast, the non-omniscient condition intentionally omits critical relational links, rendering certain spatial relations formally undecidable. This distinction allows us to evaluate not only deductive reasoning under complete information, but also the model’s ability to recognize epistemic uncertainty when the spatial structure cannot be fully determined.

\paragraph{Generation Procedure} 
All synthetic instances are generated programmatically using a Java-based spatial constructor to ensure formal consistency and reproducibility. Objects are first assigned to non-overlapping spatial regions, referred to as blocks. Each block defines a bounded subspace within the global coordinate system, and objects within a block follow strictly defined positional constraints. Pairwise spatial relation tuples (e.g., Left, Right, Above, Inside, Touch) are then derived deterministically from object coordinates, guaranteeing alignment between the underlying geometry and the textual description.

Because spatial relations are computed through explicit coordinate operations, the resulting descriptions are mathematically verifiable. Any inconsistency in model output can therefore be directly traced to reasoning errors rather than annotation noise.

The dataset comprises 80 instances in total, evenly distributed across the four generation conditions: 2D-omniscient, 2D-non-omniscient, 3D-omniscient, and 3D-non-omniscient, with 20 instances per condition. This balanced design ensures comparability across dimensional and epistemic settings while maintaining controlled experimental variation.

The generation pipeline is fully parameterized, allowing scalable expansion and fine-grained control over spatial complexity in future extensions.
\begin{table*}[t]
\centering
\caption{Hierarchy of spatial reasoning tasks in the SpatialText benchmark.}
\label{tab:task_hierarchy}
\begin{tabular}{c c c l}
\toprule
\textbf{Level} & \textbf{Subtask} & \textbf{Name} & \textbf{Targeted Ability} \\
\midrule
I. Basic Retrieval & I-I & Fact Extraction & Explicit information recall and attention \\
                   & I-II & Logical Reasoning & Simple arithmetic and set-based inference \\
\midrule
II. Static Space   & II-I & Relative Position & Direct spatial relations between objects \\
                   & II-II & Wall Mapping & Establishing a global reference frame \\
                   & II-III & Inverse Localization & Inferring locations from known anchors \\
\midrule
III. Perspective Transformation & III-I & Observer Perspective & Mental rotation under observer movement \\
                                & III-II & Coordinate Transformation & Switching between reference systems \\
\midrule
IV. Geometric Physics & IV-I & Axis and Alignment & Geometric structure and alignment reasoning \\
                     & IV-II & Visibility and Occlusion & Physical constraints and commonsense visibility \\
\midrule
V. Logical Dynamics & V-I & Path Planning & Navigation under spatial constraints \\
                   & V-II & Counterfactual Reasoning & Hypothetical spatial modification \\
                   & V-III & Logical Consistency Check & Detecting contradictions in spatial descriptions \\
                   & V-IV & Functional Inference & Inferring object affordances from layout \\
\bottomrule
\end{tabular}
\end{table*}

\begin{table*}[t]
\centering
\caption{Task distribution for human-annotated scenes across different reference frames.}
\label{tab:task_stats}
\begin{tabular}{lcccccc}
\toprule
\textbf{Task Category} & \textbf{Ego.} & \textbf{Allo.} & \textbf{Hybrid} & \textbf{Total} & \textbf{Targeted Ability} \\
\midrule
I. Basic Retrieval & 14 & 47 & 20 & 81 & Explicit information recall and attention \\
II. Static Spatial Reasoning & 28 & 56 & 46 & 130 & Direct spatial relations between objects \\
III. Perspective Transformation& 0 & 54 & 63 & 117 & Mental rotation under observer movement \\
IV. Geometric Physics & 3 & 28 & 31 & 62 & Physical constraints and visibility \\
V. Logical Dynamics & 1 & 46 & 48 & 95 & Path planning and hypothetical modification \\
\midrule
\textbf{Total} & \textbf{46} & \textbf{231} & \textbf{208} & \textbf{485} & -- \\
\bottomrule
\end{tabular}
\end{table*}

\section{Task Taxonomy and Evaluation Dimensions}
\label{sec:tasks}

To comprehensively evaluate the spatial reasoning capabilities of Large Language Models (LLMs), we design distinct Question-Answering (QA) frameworks tailored to the specific characteristics of the two data sources. For the human-annotated natural scenes, which are characterized by linguistic richness and semantic nuance, we implement a five-level hierarchical taxonomy. This multi-layered approach allows for a granular decomposition of model performance, spanning from basic retrieval to complex mental rotation and logical dynamics. In contrast, for the code-generated structured environments, our evaluation shifts toward formal rigor. Given the deterministic nature of these scenes, the QA tasks are specifically designed to probe whether models can construct a logically consistent internal representation from precise geometric descriptions and perform deductive reasoning under varying epistemic conditions.

\subsection{Tasks for Human-Annotated Data}

For the 105 real-world scenes, we construct 485 questions organized into a five-level hierarchy of increasing complexity (see Table~\ref{tab:task_hierarchy} and Table~\ref{tab:task_stats}). The primary objective of this design is to systematically probe the construction of internal spatial representations when faced with the inherent ambiguity of natural language.

Rather than treating spatial reasoning as a monolithic skill, this tiered structure allows us to pinpoint the specific boundaries of a model’s "world model." By transitioning from basic fact extraction to perspective transformation and eventually to dynamic logical synthesis, we evaluate the model’s progression from surface-level linguistic parsing to high-level spatial simulation. This approach reveals not only whether a model can recall explicit relations, but also how robustly it maintains spatial consistency when required to perform mental rotations, account for physical constraints, or reason through counterfactual modifications.

\subsection{Task for Code-Generated Data}
\label{sec:code_tasks}

The code-generated component of \textbf{SpatialText} prioritizes logical precision over linguistic variety. Because each scene is derived from an underlying geometric ground truth, our evaluation focuses on the model’s ability to "reconstruct" and "verify" spatial structures without the interference of perceptual ambiguity.

The reasoning tasks in this category are designed as a diagnostic suite. We require models to navigate the internal geometry of a scene by \textbf{inferring latent relations} between objects and \textbf{identifying specific entities} based on a conjunction of spatial constraints. Furthermore, the evaluation probes the model's structural awareness by asking it to \textbf{localize objects within hierarchical regions} (blocks) and \textbf{validate the truth value} of complex spatial propositions. Under the non-omniscient condition, this necessitates not only deductive accuracy but also the epistemic humility to recognize when a relation is formally undecidable. By distributing 12 questions per scene across these reasoning dimensions, we provide a robust metric for the model’s formal consistency and its capacity for noise-free spatial deduction.

\section{Experimental Setup}

To ensure fairness, reproducibility, and scientific rigor, we carefully control the model scale, evaluation protocol, and inference configuration across all experiments. This section details the selection of evaluated models, baseline settings, and the unified inference and prompting strategy.

\subsection{Model Selection}
\begin{table*}[t]
\centering
\caption{Performance on Human-Annotated Spatial Descriptions under Different Reference Frames.
Accuracy is omitted (denoted as ``--'') for categories with insufficient numbers of questions,
for which per-category accuracy is not statistically meaningful.}
\label{tab:human_description_results}
\resizebox{\linewidth}{!}{
\begin{tabular}{lcccccc|cccccc|cccccc|cccccc}
\toprule
 & \multicolumn{24}{c}{\textbf{Human-Annotated Descriptions}} \\
\cmidrule(lr){2-25}
Model 
& \multicolumn{6}{c}{Egocentric}
& \multicolumn{6}{c}{Allocentric}
& \multicolumn{6}{c}{Hybrid}
& \multicolumn{6}{c}{Total} \\
\cmidrule(lr){2-7} \cmidrule(lr){8-13} \cmidrule(lr){14-19} \cmidrule(lr){20-25}
 & I & II & III & IV & V & Acc.
 & I & II & III & IV & V & Acc.
 & I & II & III & IV & V & Acc.
 & I & II & III & IV & V & Acc. \\
\midrule
DeepSeek-R1-Distill-Llama-8B & 92.86& 32.14& -- & -- & -- & 52.17& 93.62& 60.71& 51.85& 42.86& 41.30& 59.31& 60.00& 32.61& 28.57& 29.03& 43.75& 36.06& 85.19& 44.62& 39.32& 37.10& 42.11& 48.66\\
OpenPangu-Embedded-7B-V1.1   & 85.71& 32.14& -- & -- & -- & 52.17& 93.62& 87.50& 70.37& 67.86& 82.61& 81.39& 90.00 & 60.87 & 61.90 & 64.52 & 60.42 & 64.42 & 91.35& 66.15& 65.81& 66.13& 71.58& 71.34\\
Qwen3-8B                     & 78.57 & 42.86 & -- & -- & -- & 56.52 & 93.62 & 89.29 & 75.93 & 60.71 & 71.74 & 80.09 & 90.00 & 60.87 & 50.79 & 54.84 & 60.42 & 59.62 & 90.12 & 69.23 & 62.39 & 59.68 & 65.26 & 69.07 \\
Gemma-3-12B-IT               & 71.43 & 35.71 & -- & -- & -- & 47.83 & 87.23 & 67.86 & 55.56 & 39.29 & 58.70 & 63.64 & 65.00 & 47.83 & 53.97 & 48.39 & 60.42 & 54.33 & 79.01 & 53.85 & 54.70 & 43.55 & 60.00 & 58.14 \\
Qwen2.5-7B                   & 78.57 & 25.00 & -- & -- & -- & 41.30 & 76.60 & 51.79 & 31.48 & 60.71 & 43.48 & 51.52 & 80.00 & 39.13 & 36.51 & 48.39 & 39.58 & 43.75 & 77.78 & 41.54 & 34.19 & 53.23 & 41.05 & 47.22 \\
Gemma-2-9B-IT                & 78.57 & 17.86 & -- & -- & -- & 39.13 & 78.72 & 67.86 & 68.52 & 53.57 & 56.52 & 66.23 & 60.00 & 52.17 & 52.38 & 67.74 & 54.17 & 55.77 & 74.07 & 51.54 & 59.83 & 61.29 & 54.74 & 59.18 \\
Mistral-7B-Instruct          & 50.00 & 25.00 & -- & -- & -- & 32.61 & 53.19 & 19.64 & 40.74 & 32.14 & 41.30 & 37.23 & 50.00 & 15.22 & 22.22 & 25.81 & 31.25 & 25.96 & 51.85 & 19.23 & 30.77 & 29.03 & 35.79 & 31.96 \\
DeepSeekv3.2                 & 92.86 & 42.86 & -- & -- & -- & 58.70 & 93.62 & 89.29 & 92.59 & 78.57 & 84.78 & 88.74 & 85.00 & 76.09 & 80.95 & 74.19 & 68.75 & 76.44 & 91.36 & 74.62 & 86.32 & 74.19 & 76.84 & 80.62 \\
\midrule
\textbf{ \#Questions}  & 14 & 28 & 0 & 3 & 1 & 46  & 47 & 56 & 54 & 28 & 46 & 231 & 20 & 46 & 63 & 31 & 48 & 208 & 81 & 130 & 117 & 62 & 95 & 485\\ 
\bottomrule
\end{tabular}
}
\end{table*}

\begin{figure*}[t]
    \centering
    % width=0.9\textwidth 表示图片宽度占正文宽度的 90%
    \includegraphics[width=1\textwidth]{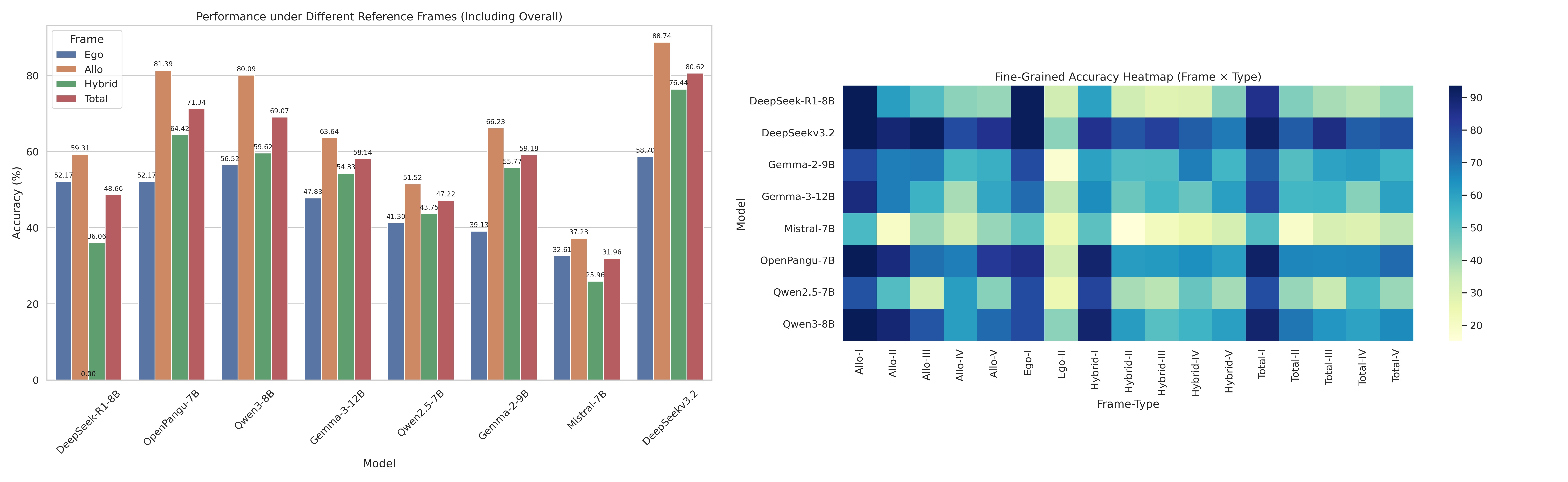} 
    
    \caption{The bar chart on the left presents the performance of different models under various description strategies, along with their overall performance. The heatmap on the right summarizes the overall results and further breaks them down by fine-grained question categories.}
    \label{fig:human_annotation_examples}
\end{figure*}

\label{subsec:model_selection}

Our evaluation focuses on \emph{lightweight large language models} in the range of 7B--14B parameters. This design choice is motivated by two considerations. 
First, models of this scale are widely adopted in practical deployment scenarios, including edge devices and resource-constrained environments. 
Second, from a cognitive perspective, this parameter regime provides an appropriate testbed for studying \emph{capability emergence}, allowing us to investigate whether complex spatial reasoning can arise from architectural or inference-level enhancements (e.g., chain-of-thought reasoning) rather than sheer parameter scaling.

To capture diverse training paradigms and inductive biases, the evaluated models are grouped into the following categories.

\paragraph{Reasoning-Enhanced Models.}
To investigate the impact of explicit reasoning supervision on spatial cognition, we include models that are designed with native multi-step reasoning capabilities. DeepSeek-R1-Distill-Llama-8B\cite{deepseekr12025} is distilled from extensive reasoning traces, allowing it to generate coherent long-chain-of-thought processes, which helps us assess how explicit supervision affects the accuracy of long-horizon spatial logic. OpenPangu-Embedded-7B-V1.1\cite{chen2025pangu}, on the other hand, is optimized for embedded and edge scenarios but retains reasoning-oriented training, providing a perspective on whether task-specific optimization can influence spatial commonsense reasoning.

\paragraph{Multimodal-Native Models.}
Although our benchmark evaluates only textual inputs, models pretrained on large-scale vision–language data offer a unique opportunity to study latent spatial representations. Qwen3-8B\cite{qwen32025} and Gemma-3-12B-IT\cite{gemma32025} fall into this category. By including these models, we can test whether visual pretraining induces internal spatial knowledge that remains accessible even when the model is restricted to text, shedding light on potential cross-modal transfer effects in spatial reasoning.

\paragraph{High-Performance Generalist Models.}
To establish strong baselines for specialized spatial reasoning, we include high-performance instruction-tuned models that excel in general tasks. Qwen2.5-7B\cite{qwen2.5} and Gemma-2-9B-IT\cite{gemma22024} represent the current state-of-the-art in open-source instruction models of this parameter scale. Their performance allows us to assess how well a general-purpose, instruction-tuned model can handle complex spatial queries without task-specific reasoning enhancements.

\paragraph{Legacy Baseline \& Large-Scale Reference Model.}

To contextualize performance gains, we include both a historical baseline and a large-scale reference. Mistral-7B-Instruct\cite{jiang2023mistral} serves as an early-generation instruction model, providing a benchmark to evaluate progress in spatial reasoning on this scale. DeepSeek-V3.2\cite{deepseekr12025} acts as a contemporary large-scale reference, illustrating how increasing parameter counts and training data can further enhance spatial reasoning capabilities beyond the primary focus range of our study.

\subsection{Implementation Details}
\label{subsec:implementation}

To ensure reproducibility and a fair comparison across all evaluated models, we standardized the inference protocol and prompt design. All experiments were conducted with deterministic decoding, setting \texttt{do\_sample=False} (equivalently, \texttt{temperature=0}), so that each model produces its most probable output sequence and reflects its confident internal reasoning.

For models with native chain-of-thought capabilities, such as \textbf{DeepSeek-R1-Distill-Llama-8B}\cite{deepseekr12025}, we preserved their intrinsic reasoning traces using standard prompts. Models without explicit CoT training, including \textbf{Qwen2.5-7B}\cite{qwen2.5} and \textbf{Mistral-7B-Instruct}\cite{jiang2023mistral}, were guided with a system-level instruction to generate step-by-step reasoning:

\begin{quote}
``Analyze the spatial description below. Before providing the final answer, please think step by step to construct a mental map of the scene, and then output the conclusion.''
\end{quote}

This approach ensures that all models output a reasoning process followed by a final answer, enabling qualitative error analysis and fair comparison of intermediate reasoning behavior.

All experiments were conducted on Huawei Ascend 910B3 NPUs, with one model allocated per card. Each NPU has 60.96GB of HBM memory, and our experiments used up to 8192 tokens per inference, sufficient to capture long-horizon reasoning and detailed chain-of-thought sequences. Whenever supported by the model API, the reasoning instruction was placed in the \texttt{system} role to guarantee consistent behavior across models.

\begin{table*}[t]
\centering
\caption{Performance on Code-Generated Structured Spatial Scenarios. 
Values with decimal points indicate overall accuracy (\%) for each setting, 
while all other entries denote the number of correctly answered questions for each question type}
\label{tab:code_generated_results}
\resizebox{\linewidth}{!}{%
\begin{tabular}{lccccc|ccccc|ccccc|ccccc}
\toprule
 & \multicolumn{20}{c}{\textbf{Code-Generated Structured Scenarios}} \\
\cmidrule(lr){2-21}
Model
& \multicolumn{5}{c}{2D\_Omniscient}
& \multicolumn{5}{c}{2D\_Non-Omniscient}
& \multicolumn{5}{c}{3D\_Omniscient}
& \multicolumn{5}{c}{3D\_Non-Omniscient} \\
\cmidrule(lr){2-6} \cmidrule(lr){7-11} \cmidrule(lr){12-16} \cmidrule(lr){17-21}
 & FR & Y/N & CO & QS & Total Acc.
 & FR & Y/N & CO & QS & Total Acc.
 & FR & Y/N & CO & QS & Total Acc.
 & FR & Y/N & CO & QS & Total Acc. \\
\midrule
DeepSeek-R1-Distill-Llama-8B & 18 & 58 & 27 & 83 & 46.5 & 36 & 52 & 48 & 84 & 55.0 & 10 & 77 & 36 & 76 & 49.8 & 35 & 42 & 39 & 74 & 48.5 \\
OpenPangu-Embedded-7B-V1.1   & 64 & 69 & 49 & 94 & 69.0 & 69 & 71 & 63 & 100 & 75.8 & 62 & 80 & 54 & 94 & 72.5 & 69 & 61 & 46 & 87 & 67.1\\
Qwen3-8B                     & 69 & 92 & 82 & 99 & 85.5 & 63 & 83 & 74 & 100 & 80.0 & 61 & 93 & 62 & 97 & 78.2 & 60 & 68 & 57 & 91 & 70.4 \\
Gemma-3-12B-IT               & 40 & 77 & 40 & 94 & 62.7 & 38 & 62 & 42 & 97 & 59.8 & 35 & 79 & 33 & 97 & 61.0 & 40 & 44 & 33 & 92 & 53.3\\
Qwen2.5-7B                   & 1 & 60 & 23 & 64 & 37.0 & 19 & 41 & 35 & 61 & 39.0 & 0 & 47 & 13 & 65 & 31.2 & 28 & 26 & 24 & 52 & 33.2 \\
Gemma-2-9B-IT                & 9 & 57 & 24 & 76 & 41.5 & 33 & 48 & 45 & 73 & 49.8 & 12 & 55 & 27 & 81 & 43.8 & 43 & 36 & 31 & 80 & 48.5 \\
Mistral-7B-Instruct          & 5 & 48 & 20 & 61 & 33.5 & 11 & 29 & 18 & 64 & 30.5 & 0 & 52 & 20 & 51 & 30.8 & 11 & 24 & 14 & 64 & 28.8 \\
\midrule
\textbf{\#Questions} 
& 100 & 100 & 100 & 100 & 400 & 100 & 100 & 100 & 100 & 400 & 100 & 100 & 100 & 100 & 400 & 100 & 100 & 100 & 92 & 392 \\
\bottomrule % 使用 bottomrule 代替末尾的 midrule 更符合规范
\end{tabular}}
\end{table*}

\begin{figure*}[t]
    \centering
    % width=0.9\textwidth 表示图片宽度占正文宽度的 90%
    \includegraphics[width=1\textwidth]{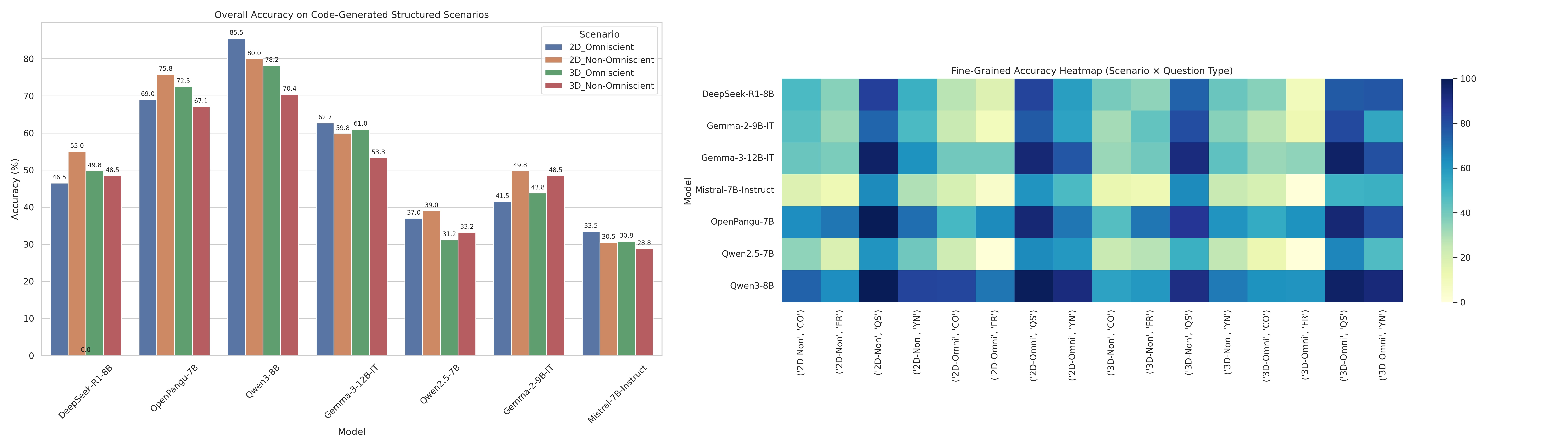} 
    
    \caption{The bar chart on the left presents the performance of different models across various evaluation dimensions and omniscient/non-omniscient settings, as well as their overall performance. The heatmap on the right summarizes the overall results and further breaks them down into fine-grained question categories.}
    \label{fig:human_annotation_examples}
\end{figure*}

% 复制并修改为 2D Non-Omniscient, 3D Omniscient, 3D Non-Omniscient

\section{Results and Analysis}

\subsection{Main results}
\label{sec:real_scene_results}

Based on overall accuracy, models can be categorized into three tiers:

\textbf{Top-tier models.} \textbf{DeepSeek-v3.2} achieves a dominant overall accuracy of 0.81, establishing the current state-of-the-art. \textbf{Qwen-Flash} (0.75) and \textbf{OpenPangu-Embedded-7B-V1.1} (0.71) follow closely, demonstrating strong spatial-semantic reasoning.

\textbf{Mid-tier models.} \textbf{Gemma-3-12b-it}, \textbf{Gemma-2-9b-it}, and \textbf{Qwen3-8B} cluster between 0.60--0.70, indicating moderate but incomplete spatial reasoning competence.

\textbf{Baseline models.} Earlier models such as \textbf{Mistral-7B-Instruct-v0.1} reach only 0.32, slightly above the random baseline (0.25). This quantifies the substantial progress made by modern LLMs in long-context spatial reasoning.

A detailed task-level analysis reveals substantial differences in model performance across cognitive dimensions, exposing specific strengths and weaknesses.

\paragraph{Type I: Basic Retrieval — Ceiling Effects from Semantic Priors.}
In factual retrieval tasks, most models (except Mistral-7B) achieve above 0.85 accuracy, with DeepSeek-v3.2 reaching 0.91.

\emph{Analysis.} High performance indicates that models can effectively leverage attention mechanisms to extract explicit spatial attributes (e.g., color, object existence). The high accuracy on \texttt{Find Block} tasks in code-generated scenarios also supports this conclusion.

\paragraph{Type III: Perspective Transformation — A Key Bottleneck.}
Perspective transformation tasks show the largest inter-model differences.

\emph{Analysis.} DeepSeek-v3.2 achieves 0.86, demonstrating robustness, while the second-best, Qwen-Flash, drops sharply to 0.70, with mid-tier models falling below 0.40. These results suggest that mental rotation and egocentric reference frame reconstruction remain challenging for most 7B--14B LLMs. DeepSeek-v3.2's superior performance likely stems from enhanced long-horizon reasoning, enabling stable maintenance of internal spatial states within the reasoning chain.

\paragraph{Type IV: Geometric Physics — Task-specific Strengths.}
In tasks involving physical constraints (e.g., occlusion, support), Qwen-Flash (0.79) slightly surpasses DeepSeek-v3.2 (0.74).

\emph{Analysis.} This suggests that the Qwen series benefits from more exposure to physically grounded or multimodal-aligned data during pretraining, improving intuitive handling of implicit physical constraints. In contrast, CoT-enhanced models such as DeepSeek-R1-Distill underperform (0.37), indicating that overly formalized reasoning may introduce noise when commonsense physical intuition is needed.

\paragraph{Type V: Logical Dynamics — Embedded Model Advantage.}
In dynamic reasoning tasks (path planning, action simulation), OpenPangu-Embedded-7B-V1.1 achieves 0.72, surpassing Qwen-Flash (0.69) and approaching DeepSeek-v3.2 (0.77).

\emph{Analysis.} Optimized for embedded or edge scenarios, this model likely benefits from training data with instruction-following or embodied control, conferring a domain adaptation advantage in navigation and action-oriented logical reasoning.

These trends are echoed in code-generated synthetic scenarios. Models maintain robust performance when object coordinates are explicit, and dimensionality effects (2D → 3D) are modest. Nevertheless, relational reasoning tasks under omniscient generation—where full context is provided—reveal notable declines, suggesting that even with complete information, models primarily rely on local relational cues rather than constructing globally coherent spatial representations.

\subsection{Cognitive Bottlenecks and Heuristic Biases}

\paragraph{The Semantic Anchor Effect.}
A critical observation across our experiments is the models' heavy reliance on semantic priors—spatial "short-cuts" derived from common real-world layouts—rather than active spatial computation. In human-annotated scenes, we identified a recurring "Bed-North" heuristic, where models consistently default to canonical orientations (e.g., assuming a bed faces North or aligns with a primary wall) regardless of the provided textual descriptions. This divergence suggests that 7B--14B models often substitute genuine spatial reasoning with probabilistic world-knowledge, failing to ground their logic in the specific, localized coordinates of the scene when they conflict with learned priors.

\paragraph{The Perspective Transformation Bottleneck.}
The transition from allocentric to egocentric reference frames represents the most significant cognitive hurdle for current LLMs. While models demonstrate a surprising resilience when moving from 2D to 3D coordinate spaces in synthetic scenes—suggesting that the raw dimensionality of data is not the primary constraint—their performance collapses during perspective transformation tasks. High-tier models like DeepSeek-v3.2 maintain a degree of stability, but mid-tier models frequently fall to near-random accuracy. This bottleneck indicates a fundamental lack of a "mental rotation" capability; the models struggle to maintain the invariance of spatial relations when the observer’s viewpoint shifts. 

\paragraph{Failure of Integrative Representation.}
Counter-intuitively, providing models with a complete relational context—referred to here as "Omniscient Generation"—often leads to a decline in performance compared to sparse, non-omniscient scenarios. This "Paradox of Omniscience" reveals that current LLMs primarily employ a local reasoning strategy, focusing on immediate, pairwise object relations rather than constructing a globally coherent spatial representation. In dense synthetic environments where the relational graph is fully articulated, models become overwhelmed by the information density, leading to logical contradictions and failures in global consistency. This suggests that even with "all the answers" provided in the prompt, models lack the integrative architecture required to synthesize a unified spatial map. Consequently, as scene complexity increases, the models' reliance on local cues becomes a liability, preventing them from resolving multi-object spatial hierarchies effectively.

\subsection{Summary}

The systematic evaluation of state-of-the-art LLMs on SpatialText reveals a nuanced landscape of spatial cognition. While models demonstrate near-perfect accuracy in Type I (Basic Retrieval) tasks, indicating robust information extraction and attention mechanisms, a significant performance "cliff" is observed as tasks move toward Type III (Perspective Transformation) and Type V (Logical Dynamics).

A pivotal discovery in our analysis is the "Orientation-Heuristic Gap." We observed a persistent failure pattern in egocentric reasoning, most notably the "Bed-North" hallucination: when a person is described as lying supine with their head toward the North, models consistently fail to correctly identify relative directions (e.g., left/right), often defaulting to high-frequency linguistic associations (pairing "Left" with "West") rather than performing the necessary geometric rotation. This suggests that current LLMs, even those with strong reasoning traces like DeepSeek-R1, still rely heavily on statistical linguistic heuristics—spatial patterns frequently co-occurring in text—rather than constructing a coherent, verifiable internal mental map.

Furthermore, the results from code-generated scenarios confirm that while models can handle increased dimensionality (3D), their reasoning remains localized. The decline in performance under Omniscient settings for complex relational tasks indicates that models struggle with global consistency; they can parse individual relations but fail to integrate them into a unified spatial manifold. These findings highlight that the bottleneck in machine spatial intelligence is not the complexity of the data, but the lack of an intrinsic mechanism for embodied perspective-taking and global topological verification.

\section{Conclusion}

In this paper, we introduced SpatialText, a pioneering text-only benchmark designed to probe the cognitive boundaries of large language models in 3D spatial reasoning. By employing a dual-source data strategy—grounding naturalistic human descriptions in real-world scenes while providing a rational skeleton through code-generated structured environments—we created a rigorous framework that isolates spatial cognition from visual perception.

Our comprehensive evaluation of representative LLMs (7B to 671B parameters) yields three major insights. First, there is a clear decoupling between linguistic fluency and spatial grounding; models can describe a scene without truly "understanding" its geometric constraints. Second, perspective-taking remains the most significant hurdle, where models succumb to textual priors instead of mental rotation. Third, while large-scale models like DeepSeek-V3 show emerging signs of spatial consistency, the "spatial reasoning" of smaller models is largely a process of sophisticated pattern matching.

\bibliography{ecai}
\end{document}